\documentclass[10pt,twocolumn,letterpaper]{article}

\usepackage[normalem]{ulem}

\usepackage{wacv}              

\usepackage{adjustbox}
\usepackage{graphicx}
\usepackage{comment}

\usepackage{multirow}
\usepackage{arydshln}

\definecolor{wacvblue}{rgb}{0.21,0.49,0.74}
\usepackage[pagebackref,breaklinks,colorlinks,allcolors=wacvblue]{hyperref}

\def\wacvPaperID{66} 
\def\confName{WACV}
\def\confYear{2027}
\usepackage[accsupp]{axessibility} 

\title{SVD-Prune: Spectral Vision Token Pruning under Extreme Compression}

\author{
Yvon Apedo$^{1}$,
Martyna Poreba$^{2}$,
Michal Szczepanski$^{2}$,
Samia Bouchafa$^{1}$
\\[0.5em]
 $^{1}$Université Paris-Saclay, Univ Evry, IBISC, 91020, Evry-Courcouronnes, France
\\
$^{2}$Université Paris-Saclay, CEA, List, F-91120 Palaiseau, France
\\
}

\begin{document}
\maketitle

\begin{abstract}

Vision-Language Models (VLMs) and Vision-Language-Action (VLA) models process hundreds of visual tokens, making visual representations a major computational bottleneck. 
We argue that visual token pruning should be formulated not as identifying individually important tokens, but as preserving the dominant low-rank spectral structure of the visual representation. 
Based on this formulation, we introduce SVD-Prune, a training-free framework that leverages Singular Value Decomposition (SVD) and statistical leverage scores for globally informed token selection. SVD-Prune is architecture-agnostic and can be applied at different stages of the multimodal pipeline, including encoder-side, decoder-side, and hybrid configurations Extensive experiments across VLM and VLA benchmarks demonstrate strong robustness under extreme compression. SVD-Prune reduces the visual token. SVD-Prune cuts the visual token budget by up to 99.3\%, while, at a 16-token budget, encoder-side pruning lowers inference FLOPs by 82.4\% and achieves a 4.18$\times$ end-to-end speedup, including pruning overhead. 
Code: \url{https://github.com/anoncvlab/svdPrune} 
\end{abstract}

\section{INTRODUCTION}
\label{sec:intro}

Multimodal Large Language Models (MLLMs), particularly Vision-Language Models (VLMs) and their action-oriented extensions such as Vision-Language-Action (VLA) models, have emerged as a dominant paradigm for multimodal reasoning and embodied intelligence. By combining visual perception with language understanding, these models enable a broad range of capabilities, from visual question answering to robotic manipulation. Despite their success, visual representations remain a major computational bottleneck. To address this inefficiency, token pruning has become a central direction for efficient multimodal inference. By removing redundant visual tokens, recent methods substantially reduce latency and computational cost. Many existing approaches derive token-wise selection scores from attention statistics, token norms, cross-modal relevance, or related importance signals. Although effective at moderate compression ratios, these criteria do not explicitly measure each token's contribution to the global structure of the full token set. This limitation becomes particularly relevant under aggressive compression, where token-wise scores may fail to reflect the collective structure of the visual representation.

We argue that visual token pruning can instead be viewed as a low-rank subspace preservation problem: the objective is to retain a compact subset of tokens that remains representative of the dominant spectral structure of the complete visual representation. Based on this formulation, we introduce SVD-Prune, a training-free, attention-independent framework that leverages spectral decomposition and statistical leverage scores for globally informed token selection. Our contributions are threefold:

\begin{itemize}
\item We reformulate visual token pruning as a low-rank subspace preservation problem, providing a globally informed alternative to token selection criteria derived from local interactions; 

\item  We introduce SVD-Prune, a spectral token pruning framework applicable at the encoder side, decoder side, or in a hybrid configuration;

\item Through a unified evaluation across VLMs and VLAs, we demonstrate the robustness of SVD-Prune across multimodal reasoning and embodied decision-making tasks, including under extreme compression, while substantially reducing visual tokens and inference cost. 

\end{itemize}

\section{RELATED WORK}
\label{sec:related_work}

VLMs typically encode images into hundreds of visual tokens, substantially increasing sequence-dependent computation during multimodal inference. This effect can be further amplified in VLA architectures that employ multiple visual encoders, leading to larger visual token representations~\cite{kim2024openvla, li2026SemanticVLA}. Existing visual token reduction methods for VLMs can be broadly categorized according to where pruning is applied. Encoder-side approaches reduce tokens within the vision backbone or immediately after visual encoding. ToMe~\cite{bolya2022token} merges redundant tokens directly inside the Vision Transformer, while FasterVLM~\cite{zhang2024cls}, HiRED~\cite{arif2025hired}, DivPrune~\cite{alvar2025divprune}, VisionZip~\cite{yang2025visionzip}, and TRIM~\cite{song2025trim} operate on encoder outputs. In particular, FasterVLM and HiRED rely on attention-derived importance scores, DivPrune explicitly promotes diversity among retained tokens, VisionZip combines token selection with similarity-based merging, and TRIM exploits cross-modal similarity to identify visually relevant regions. Decoder-side approaches prune visual tokens within the language model during multimodal reasoning. FastV~\cite{chen2024fastv} removes tokens after early decoder layers using attention statistics, FitPrune~\cite{ye2025fitprune} introduces handcrafted pruning schedules, SparseVLM~\cite{zhang2025sparsevlm} performs adaptive text-aware token sparsification with token recycling, and PDrop~\cite{wang2025PDrop} progressively increases pruning ratios across decoder depth. Beyond these two categories, hybrid approaches perform token reduction at multiple stages of the vision-language pipeline. IVTP~\cite{huang2024ivtp} first removes visually redundant tokens in the vision encoder using attention-rollout-based importance scores, and subsequently prunes instruction-irrelevant visual tokens within the LLM through CLIP-guided text-aware token selection.

Recent work has extended visual token reduction to VLA models, initially by adapting pruning strategies developed for VLMs. In particular, FastV~\cite{chen2024fastv} and SparseVLM~\cite{zhang2025sparsevlm} have been directly applied to VLA architectures, leveraging respectively attention-based saliency scores and text-aware token sparsification. However, these methods do not explicitly incorporate action-specific or temporal signals. EfficientVLA~\cite{yang2026efficientvla} addresses this limitation by performing visual token selection on encoder outputs using a combination of instruction-aware relevance scores and feature-diversity measures. More recently, VLA-specific approaches such as SP-VLA~\cite{li2025sp}, SpecPrune-VLA~\cite{wang2025specprunevla}, VLA-Pruner~\cite{liu2025vlapruner}, and ADP~\cite{pei2025action} incorporate action-aware or temporal information into the pruning process, exploiting the sequential nature of robotic interaction to better preserve action-critical visual cues. LightVLA~\cite{lightvla} introduces a differentiable token pruning framework that learns task-driven visual token importance through dynamic queries to adaptively prune visual tokens. Notably, most VLA-specific methods operate at the encoder level, reducing visual computation without modifying autoregressive action generation.

\section{METHODOLOGY}
\label{sec:methodology}

\begin{figure*}[!t]
    \centering
    \includegraphics[width=1.0\linewidth]{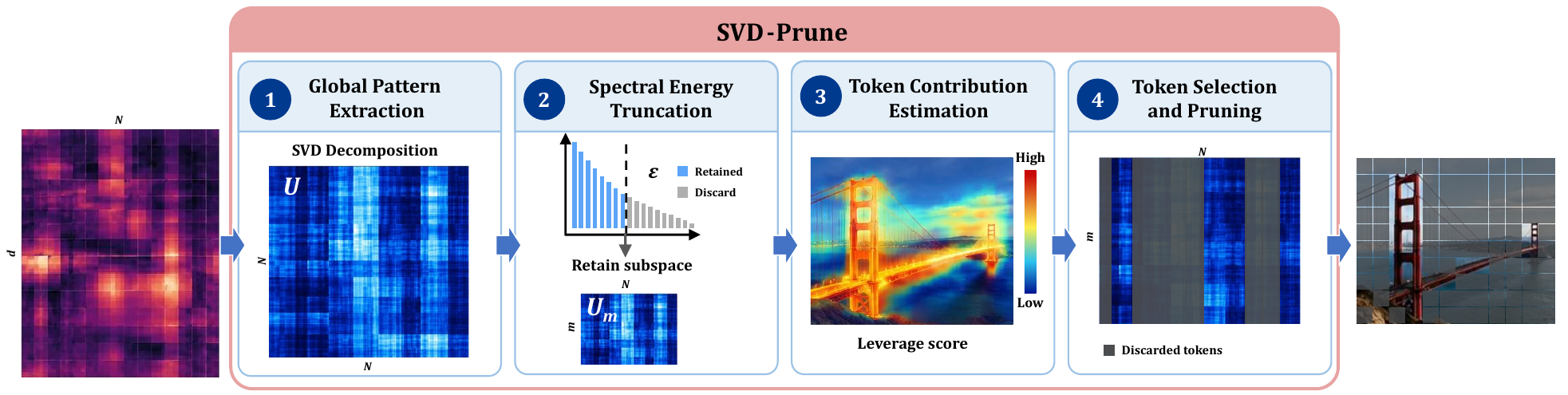}
    \caption{Overview of SVD-Prune. Visual representations are decomposed using SVD to identify a dominant spectral subspace, and tokens are ranked using statistical leverage scores to retain a compact representative subset. $N$ denotes the number of tokens, $d$ the feature dimension, and $m$ the truncation rank.
    }
    \label{fig:svd}
\end{figure*}

Rather than estimating token importance from local interactions, we formulate visual token pruning as the problem of preserving the dominant low-rank structure underlying the full set of visual tokens. To this end, we propose SVD-Prune, a training-free and attention-independent framework that identifies a dominant low-rank spectral subspace and ranks tokens according to their leverage scores within this subspace. This globally informed criterion favors a compact subset of tokens that is representative of the dominant low-rank structure while remaining applicable across encoder-side, decoder-side, and hybrid pruning configurations. Figure~\ref{fig:svd} provides an overview of SVD-Prune. The pipeline consists of four main stages: (1) global pattern extraction, (2) spectral energy truncation, (3) token contribution estimation via leverage scores, and (4) token selection and pruning.

\subsection{Global Pattern Extraction}
\label{ssec:extraction}

At a given pruning location, let $\mathbf{F} \in \mathbb{R}^{N \times d}$ denote the matrix of visual token representations, where $N$ is the number of visual tokens and $d$ their feature dimension. Depending on the pruning configuration, $\mathbf{F}$ may correspond to intermediate encoder features, encoder outputs, or visual-token representations within the language model. Visual token representations often exhibit strong correlations across tokens, motivating the characterization of their dominant low-rank structure. 
We apply Singular Value Decomposition (SVD) to the feature matrix

\begin{equation}
\mathbf{F} = \mathbf{U} \boldsymbol{\Sigma} \mathbf{V}^\top,
\label{eq:decomposition}
\end{equation}

where $\mathbf{U}\in\mathbb{R}^{N\times N}$ contains the left singular vectors, $\boldsymbol{\Sigma}\in\mathbb{R}^{N\times d}$ contains the singular values $s_1 \geq s_2 \geq \cdots \geq s_{\min(N,d)} \geq 0$ on its diagonal, and $\mathbf{V}\in\mathbb{R}^{d\times d}$ contains the corresponding right singular vectors. 
Since the feature matrix is not centered prior to decomposition, the singular values characterize the spectral energy of the representation rather than variance in the PCA sense.

\subsection{Spectral Energy Truncation}
\label{ssec:var_pattern}

After decomposition, we identify the dominant spectral components by analyzing the energy associated with the singular values. Specifically, we compute the normalized spectral energy of each component as 
\begin{equation}
e_i = \frac{s_i^2}{\sum_{j=1}^{\min(N,d)} s_j^2},
\label{eq:variance}
\end{equation}
where $s_i$ denotes the $i$-th singular value. The cumulative retained energy up to rank $m$ is then defined as
\begin{equation}
c_m = \sum_{i=1}^{m} e_i.
\label{eq:cumulative}
\end{equation}

We select the smallest rank $m$ such that $c_m \geq \varepsilon$, where $\varepsilon \in (0,1]$ controls the fraction of retained spectral energy and thereby determines the rank of the truncated subspace used for token scoring. According to the Eckart--Young theorem, the resulting rank-$m$ approximation is optimal with respect to Frobenius reconstruction error among all rank-$m$ approximations of $\mathbf{F}$. Importantly, $\varepsilon$ does not directly determine the pruning ratio. The final compression level is instead controlled independently by the token budget $k$ during the selection stage.

\subsection{Token Contribution Estimation}
\label{ssec:leverage}

Having identified the dominant spectral subspace, we quantify each token's contribution using statistical leverage scores~\cite{mahoney2009cur}. In our setting, leverage scores provide a subspace-aware measure of each visual token's influence on the retained low-rank structure. For each token $t \in \{1,\ldots,N\}$, the standard rank-$m$ leverage score is defined as

\begin{equation}
h_t = \sum_{j=1}^{m}(U_{t,j})^2
= \bigl\lVert \mathbf{U}_{m}[t,:] \bigr\rVert_2^2,
\label{eq:leverage_score}
\end{equation}

where $h_t$ denotes the standard rank-$m$ statistical leverage score of token $t$. Since the columns of $\mathbf{U}_m$ are orthonormal, $\sum_{t=1}^{N} h_t = m$. We use the normalized score

\begin{equation}
\ell_t = \frac{h_t}{m},
\end{equation}

such that $\sum_{t=1}^{N} \ell_t = 1$. This normalization does not affect token ranking. Tokens with higher leverage scores exert greater influence on the retained low-rank subspace and are therefore favored during selection, whereas low-leverage tokens have lower influence on this subspace. 

\subsection{Token Selection and Pruning}
\label{ssec:selection}

Tokens are ranked according to their leverage scores, and the top-$k$ tokens are retained, where $k$ denotes the target token budget. The selected token indices are then restored to their original 
sequence order to preserve positional consistency and maintain compatibility with downstream attention mechanisms. This deterministic top-$k$ selection serves as a practical heuristic for identifying representative tokens. While leverage scores provide a subspace-aware ranking criterion, selecting the top-$k$ tokens does not guarantee an optimal subset for downstream task performance. Instead, it favors tokens with the largest influence on the retained spectral subspace. Importantly, leverage scores measure token influence with respect to the retained subspace rather than pairwise diversity. Consequently, deterministic top-k selection does not explicitly prevent multiple similar high-leverage tokens from being retained; enforcing explicit diversity would require an additional subset-selection criterion.

\section{EXPERIMENTS AND ANALYSIS}
\label{sec:experiments_analysis}

\subsection{Experimental setup}
\label{ssec:experimental_settings}

We evaluate SVD-Prune on both VLMs and VLAs. For VLM experiments, we use LLaVA-1.5-7B~\cite{liu2023llava} as our primary benchmark and additionally evaluate QwenVL~\cite{bai2023qwenvl} to assess generalization across different VLM architectures. Following the standard LLaVA-1.5 configuration, input images are resized to $336\times336$, yielding 576 visual tokens. QwenVL processes images at a higher resolution of $448\times448$, resulting in a larger visual token budget of 1024 tokens. Additional experiment results on LLaVA-1.5-13B 
are reported in the supplementary materials. Performance is evaluated
on a diverse collection of multimodal benchmarks covering visual reasoning, object perception, hallucination robustness, scientific reasoning, and text-centric visual understanding~\cite{lu2022sqa, gurari2018vizwiz, li2023pope, hudson2019gqa, singh2019towards, fu2023mme}. To assess embodied decision-making capabilities, we additionally consider OpenVLA~\cite{kim2024openvla} on the LIBERO-Spatial, LIBERO-Goal, LIBERO-Object, and LIBERO-10 benchmarks~\cite{liu2023libero}. OpenVLA represents each visual observation using 256 visual tokens. Performance is measured in terms of task success rate and inference efficiency. For OpenVLA, we evaluate all 10 tasks in each LIBERO suite with 50 trials per task, resulting in 500 trials per suite, using a fixed random seed of 7. All experiments are conducted using PyTorch on a single NVIDIA RTX 3080 16\,GPU with an Intel Core i7-11800H CPU.

\subsection{Pruning Across the VLM}
\label{sec:VLM_results}

\subsubsection{Pruning Location Analysis}
Table~\ref{tab:perf_enc_LLava7B} summarizes the performance of SVD-Prune across pruning configurations and token budgets. We extend the comparison to extreme compression by evaluating competing methods at token budgets not considered in the original works. Owing to its architecture-agnostic formulation, SVD-Prune can be applied at different stages of the multimodal pipeline, including encoder-side, decoder-side, and hybrid configurations that combine both pruning stages.

\begin{table*}[!t]
\centering
\footnotesize
\setlength{\tabcolsep}{9pt} 

\begin{tabular}{@{} l l c c c c c c @{}}


\toprule
\textbf{Methods} & \textbf{Location} & \textbf{SQA}~\cite{lu2022sqa} & \textbf{VizWiz}~\cite{gurari2018vizwiz} & \textbf{POPE}~\cite{li2023pope} & \textbf{GQA}~\cite{hudson2019gqa} & \textbf{TextVQA}~\cite{singh2019towards} & \textbf{MME}~\cite{fu2023mme} \\
\midrule
Vanilla & -- & 70.43 & 50.00 & 86.96 & 61.90 & 58.20 & 1862.00 \\
\midrule
\multicolumn{8}{c}{\textit{Retain 192 Tokens (33.3\%)}} \\
\midrule
FastV~\cite{chen2024fastv}  & Decoder & 69.10 & 50.80 & 64.80 & 52.88 & 52.50 & 1605.00 \\
PDrop~\cite{wang2025PDrop}  & Decoder & \underline{70.02} & \underline{51.10} & 82.30 & 57.30 & 56.50 & 1766.00 \\

SparseVLM~\cite{zhang2025sparsevlm} & Decoder &  69.10 & 52.90 & 83.60 & 57.60 & 56.10 & 1721.00 \\
VisionZip~\cite{yang2025visionzip} & Encoder &  68.90 & 54.10 & 85.30 & 59.30 & \textbf{57.30} & 1782.60 \\
\hdashline
SVD-Prune (ours) & Encoder & \textbf{70.08} & \textbf{54.78} & \textbf{87.75} & \underline{59.88} & \underline{57.24} & \underline{1788.00} \\
 & Decoder & 69.87 & 47.77 & \underline{87.34} & \textbf{59.94} & 32.75/55.89$^\ast$ & \textbf{1801.53} \\
\midrule
\multicolumn{8}{c}{\textit{Retain 128 Tokens (22.2\%)}} \\
\midrule
FastV~\cite{chen2024fastv}  & Decoder & 68.60 & 51.30 & 53.40 & 49.60 & 50.60 & 1490.00 \\
PDrop~\cite{wang2025PDrop}  & Decoder & \textbf{69.90} & 51.00 & 82.30 & 57.10 & \textbf{56.60} & 1664.00 \\
SparseVLM~\cite{zhang2025sparsevlm} & Decoder &  67.10 & 51.80 & 80.50 & 56.00 & 54.90 & 1696.00 \\
VisionZip~\cite{yang2025visionzip} & Encoder & 68.10 & \underline{52.90} & 83.20 & 57.60 & 55.80 & \textbf{1761.70} \\
\hdashline
SVD-Prune (ours) & Encoder & \underline{69.84} & \textbf{55.39} & \underline{86.73} & \underline{58.70} &  \underline{56.14} & 1674.00 \\
 & Decoder & 69.51 & 48.60 & \textbf{87.10} & \textbf{58.83} & 33.53/54.88$^\ast$ & \underline{1728.74} \\
\midrule
\multicolumn{8}{c}{\textit{Retain 64 Tokens (11.1\%)}} \\
\midrule
FastV~\cite{chen2024fastv}  & Decoder & 51.10 & 50.80 & 38.20 & 46.10 & 47.80 & 1255.00 \\
PDrop~\cite{wang2025PDrop}  & Decoder & 69.20 & 50.70 & 55.90 & 47.50 & 50.60 & 1092.00 \\
SparseVLM~\cite{zhang2025sparsevlm} & Decoder & 69.70 & 50.40 & 77.50 & 53.70 & 53.40 & 1559.00 \\
VisionZip~\cite{yang2025visionzip} & Encoder &  69.00 & 53.70 & 77.00 & \underline{54.10} & \underline{55.05} & \textbf{1590.00} \\
\hdashline
SVD-Prune (ours) & Encoder & \textbf{70.12} & \textbf{54.73} & \textbf{83.87} & 53.77 & \textbf{55.14} & \underline{1575.00} \\
 & Decoder & \underline{70.05} & 47.27 & \underline{82.94} & \textbf{54.31} & 29.22/53.59$^\ast$ & 1568.71 \\
\midrule
\multicolumn{8}{c}{\textit{Retain 32 Tokens (5.6\%)}} \\
\midrule
FastV~\cite{chen2024fastv}  & Decoder & 42.60 & 51.70 & 32.50 & 41.50 & 42.50 & 884.60 \\
SparseVLM~\cite{zhang2025sparsevlm} & Decoder & 57.30 & 51.90 & 67.90 & 48.30 & 46.10 & 1046.70 \\
VisionZip~\cite{yang2025visionzip} & Encoder & 68.80 & 52.90 & 68.70 & 51.80 & 53.10 & 1247.40 \\
\hdashline
SVD-Prune (ours) & Encoder & \textbf{70.01} & \underline{53.59} & 79.34 & \underline{53.52} & \underline{54.81} & 1436.00 \\
 & Decoder & 69.49 & 47.83 & \underline{79.51} & 49.38 & 24.02/52.95$^\ast$ & \underline{1455.90} \\
 & Hybrid  & \underline{69.84} & \textbf{54.13} & \textbf{83.49} & \textbf{53.89} & 33.49/\textbf{55.11}$^\ast$ & \textbf{1563.82} \\
\midrule
 
\multicolumn{8}{c}{\textit{Retain 16 Tokens (2.8\%)}} \\
\midrule
FastV~\cite{chen2024fastv}& Decoder & 42.12 & 50.67 & 31.45 & 41.28 & 40.33 & 877.16\\
SparseVLM~\cite{zhang2025sparsevlm} & Decoder &  69.04 & 50.87 & 63.15 & 43.89 & 53.97 & 1205.73 \\
VisionZip~\cite{yang2025visionzip} & Encoder & 69.02 & 51.29 & 63.79 & 44.17 & 53.33 & 1209.25 \\
\hdashline
SVD-Prune (ours) & Encoder & \underline{69.50} & \underline{52.60} & \underline{73.90} & \textbf{53.00} & \underline{54.00} & 1281.00 \\
 & Decoder & 69.16 & 46.66 & 66.28 & 43.85 & 19.31/51.66$^\ast$  & \underline{1284.91} \\
 & Hybrid  & \textbf{70.01} & \textbf{52.92}    & \textbf{77.33} & \underline{49.02} & 26.74/\textbf{54.75}$^\ast$ & \textbf{1364.13} \\

\midrule
\multicolumn{8}{c}{\textit{Retain 8 Tokens (1.4\%)}} \\
\midrule
FastV~\cite{chen2024fastv}& Decoder &  41.85 & 47.02 & 31.00 & 38.06 & 10.99 & 859.32 \\
SparseVLM~\cite{zhang2025sparsevlm} & Decoder &  67.54 & 48.68 & 62.87 & 41.27 & \textbf{23.81} & 990.52 \\
VisionZip~\cite{yang2025visionzip} & Encoder & 67.60 & 48.66 & 54.47 & 38.87 & 13.14 & 987.63 \\
\hdashline
SVD-Prune (ours) & Encoder & \underline{69.23} & \underline{49.82} & \textbf{68.39} & \underline{41.66} & 16.64 & \underline{1130.72} \\
 & Decoder & 68.22 & 45.88    & 58.14 & 40.38 & 13.68/15.10$^\ast$   & 1077.57 \\
 & Hybrid  & \textbf{69.46} & \textbf{51.45}  & \underline{68.34} & \textbf{44.79} & 21.55/\underline{22.07}$^\ast$  & \textbf{1261.05} \\
\midrule
\multicolumn{8}{c}{\textit{Retain 4 Tokens (0.7\%)}} \\
\midrule
FastV~\cite{chen2024fastv}& Decoder &  40.01 & 45.98 & 30.98 & 37.81 & 8.09 & 849.32 \\
SparseVLM~\cite{zhang2025sparsevlm} & Decoder &  67.93 & 48.24 & \underline{60.94} & \underline{40.93} & \textbf{19.09} & 950.88 \\
VisionZip~\cite{yang2025visionzip} & Encoder & 67.51 & 47.96 & 50.81 & 36.60 & 8.51 & 930.28 \\
\hdashline
SVD-Prune (ours) & Encoder & \textbf{68.69} & \underline{48.67} & \textbf{61.46} & 39.71 & 13.53 & \underline{1054.81} \\
 & Decoder & 67.86 & 47.20   & 54.02 & 38.47 & 8.63/12.46$^\ast$ & 1030.25 \\
 & Hybrid  & \underline{67.96} & \textbf{49.85} & 59.99 & \textbf{41.14} & 15.85/\underline{16.85}$^\ast$   & \textbf{1128.01} \\

\bottomrule
\end{tabular}
\caption{Response quality of LLaVA-1.5-7B with SVD-Prune across multimodal benchmarks and token budgets. Decoder-side pruning is applied after the 8th layer, while hybrid pruning first reduces visual tokens to 64 after the encoder, followed by a second pruning stage after the 8th decoder layer. Best and second-best results are marked in \textbf{bold} and \underline{underlined}, respectively. Results marked with $^\ast$ correspond to SVD-Prune applied after the 12th decoder layer.}

\label{tab:perf_enc_LLava7B}
\end{table*}

\begin{figure*}[!ht]
    \centering
    \includegraphics[width=1.0\linewidth]{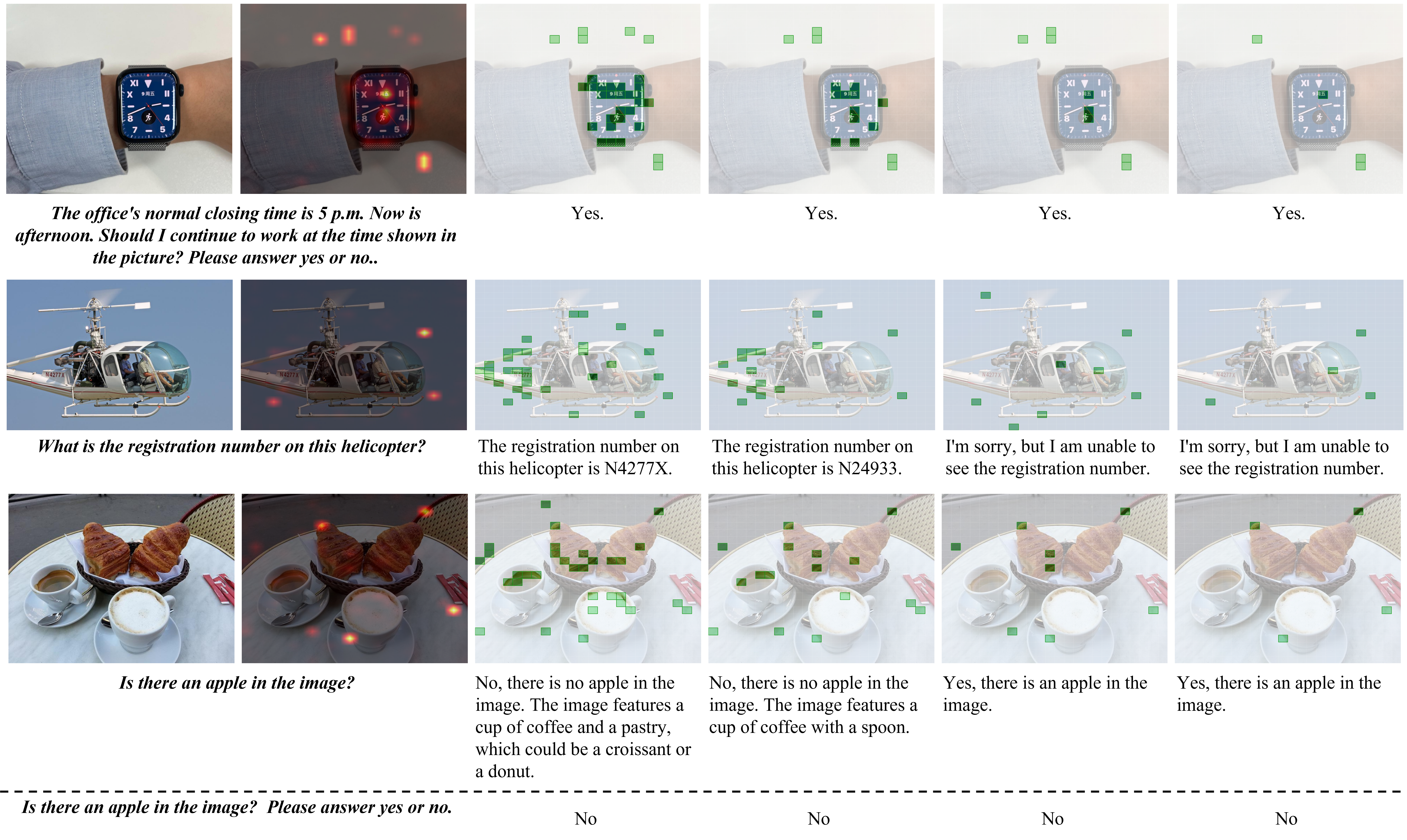}
    \caption{Qualitative visualization of SVD-Prune on representative examples from MME (top), TextVQA (middle), and POPE (bottom). From left to right: input image with the associated prompt, token leverage-score heatmap, and encoder-side pruning results retaining 32, 16, 8, and 4 visual tokens, respectively. Green markers indicate the spatial locations of the selected tokens.}
    \label{fig:vis_VLM}
\end{figure*}

\paragraph{Encoder-Side.}
We first analyze encoder-side pruning, where token selection is applied immediately after the vision encoder. At moderate token budgets (192 and 128 retained tokens), SVD-Prune remains close to the unpruned baseline while retaining only 22--33\% of the original visual tokens. 
Under more aggressive compression, performance degradation becomes more pronounced across all methods. 
At 32 and 16 tokens, encoder-side SVD-Prune matches or outperforms the competing methods across all six benchmarks. compression. Even at 8 and 4 tokens, SVD-Prune achieves the best results on five and four benchmarks, respectively, with the main exceptions being TextVQA at both budgets and GQA at 4 tokens. These results also reveal increasing task dependence under extreme compression. Reasoning-oriented benchmarks such as SQA remain highly robust, with encoder-side pruning retaining 68.69 accuracy with only 4 visual tokens, less than two points below the unpruned model. In contrast, tasks requiring fine-grained spatial perception or localized text recognition, particularly TextVQA, become significantly more sensitive to aggressive pruning. Interestingly, as illustrated in Figure~\ref{fig:vis_VLM}, prediction quality under aggressive pruning also depends on the structure of the output space. For visually similar queries, binary prompts constrained to a yes/no answer remain significantly more robust than open-ended responses. We hypothesize that restricting the output space reduces generative uncertainty and limits hallucination, making the model less sensitive to information loss caused by aggressive token pruning.

\paragraph{Decoder-side.}
We next evaluate decoder-side pruning after the 8th LLM layer, selected as a practical default across benchmarks to balance multimodal contextualization and computational savings (see Supplementary Material). At moderate token budgets, decoder-side SVD-Prune remains competitive with the strongest baselines on SQA, POPE, GQA, and MME, despite selecting visual tokens without using the text prompt. This contrasts with text-guided pruning approaches such as SparseVLM, which explicitly exploit prompt information to identify relevant visual tokens. However, a pronounced gap emerges on TextVQA, where prompt-guided selection is particularly beneficial. To investigate this behavior, we additionally apply SVD-Prune after the 12th decoder layer on TextVQA. Delaying pruning substantially recovers performance down to 16 retained tokens, suggesting that additional decoder processing helps consolidate localized textual information before query-agnostic pruning. At 8 and 4 tokens, however, the gains remain limited, indicating that the extreme token bottleneck becomes dominant. 

\paragraph{Hybrid mode.}

Hybrid pruning first reduces the visual sequence to 64 tokens after the vision encoder, followed by a second pruning stage after the 8th decoder layer. Under extreme compression (32 tokens and below), this progressive strategy becomes the strongest SVD-Prune configuration on most benchmarks, consistently improving over single-stage decoder pruning and frequently outperforming encoder-side pruning. This suggests that distributing pruning across the multimodal pipeline is more robust than performing the entire reduction at a single stage. TextVQA remains an exception, highlighting the importance of pruning location for tasks relying on fine-grained textual information. 

\subsubsection{Efficiency Analysis}

Table~\ref{tab:pruning-perf} compares the efficiency of SVD-Prune on LLaVA-1.5-7B with existing visual token pruning methods under a common final budget of 16 tokens. Encoder-side SVD-Prune reduces the computational cost from 3.450\,T to 0.606\,T FLOPs and latency from 930.00\,ms to 222.56\,ms, corresponding to a \(4.18\times\) speedup. Intra-encoder pruning further lowers the cost to 0.501\,T FLOPs and the latency to 214.27\,ms. Both configurations outperform FastV and SparseVLM in latency, while VisionZip remains faster (184.52 vs. 222.56\,ms at 16 tokens). However, SVD-Prune achieves higher accuracy than VisionZip across all six benchmarks at this budget, with particularly notable gains on GQA (+8.83 points) and POPE (+10.11 points), highlighting an accuracy--latency trade-off. The hybrid mode reaches the lowest overall cost, at 0.476\,T FLOPs, but not the lowest latency, illustrating that FLOPs reduction does not directly translate into proportional runtime gains.

Table~\ref{tab:latency_breakdown_compact} further analyzes the runtime overhead of SVD-Prune. In encoder mode, SVD decomposition, leverage scoring, and token selection add 24.73\,ms in total, while the LLM latency drops from 907.12\,ms to 174.95\,ms. Decoder-side pruning incurs a larger SVD latency (43.50\,ms) and provides limited acceleration because pruning is applied only after layer 8. The hybrid mode performs two pruning operations, with a combined overhead of 30.43\,ms. Despite this measurable latency overhead, the arithmetic cost of SVD-Prune remains small, ranging from 1.740 to 5.181\,GFLOPs, i.e., below 0.4\% of the corresponding total model cost.

\begin{table}[ht]
\centering
\small
\begin{tabular}{llccc}
\toprule
\textbf{Method} & \textbf{Location} & \textbf{Latency} & \textbf{FPS} & \textbf{FLOPs} \\
 & & [ms] & & [T] \\
\midrule
Vanilla & -- & 930.00 & 1.06 & 3.450 \\
\midrule
FastV~\cite{chen2024fastv} 
    & Decoder & 332.14 & 3.01 & 1.232 \\
SparseVLM~\cite{zhang2025sparsevlm} 
    & Decoder & 372.00 & 2.69 & 1.380 \\
VisionZip~\cite{yang2025visionzip} 
    & Encoder & 184.52 & 5.42 & 0.604 \\
\hdashline
 
SVD-Prune (ours)    & Encoder & 222.56 & 4.33 & 0.606 \\
 
    & Decoder & 886.00 & 1.13 & 1.337 \\
 
    & Hybrid & 487.00 & 2.05 & 0.476 \\
\bottomrule
\end{tabular}

\caption{Performance comparison on an RTX 3080 16GB GPU. All pruning methods are evaluated using the same final visual token budget of 16 tokens; the vanilla model uses its original 576 visual tokens.}
\label{tab:pruning-perf}
\end{table}

\begin{table}[!ht]
\centering
\small
\setlength{\tabcolsep}{6pt}
\begin{tabular}{lccc}
\toprule
\textbf{Component}
&\multicolumn{3}{c}{ \textbf{Pruning Location}} \\
 & \textbf{Encoder} & \textbf{Decoder} & \textbf{Hybrid} \\
\midrule

\multicolumn{4}{c}{\textit{Latency [ms]}} \\
Encoder
& 22.88 & 22.88 & 22.88 \\
\hdashline
\multicolumn{4}{c}{\textit{SVD-Prune overhead}} \\
SVD 
& 23.94 & 43.50 & 23.94 + 5.45 \\
Leverage Scoring  
& 0.53  & 0.53  & 0.53 + 0.15 \\
Token Selection   
& 0.26  & 0.26  & 0.26 + 0.10 \\
\hdashline
Decoder 
& 174.95 & 818.83 & 433.69 \\
\midrule
\multicolumn{4}{c}{\textit{SVD-Prune computational overhead}} \\
GFLOPs 
& 1.740 & 5.181 & 1.740 + 0.0680 \\
\bottomrule

\end{tabular}

\caption{Latency breakdown and computational overhead of different SVD-Prune deployment modes with a 16-token budget. Decoder-side pruning is applied after LLM layer 8 in both the decoder and hybrid modes. For the hybrid mode, latency and FLOPs are reported as encoder-side + decoder-side SVD-Prune costs.}
\label{tab:latency_breakdown_compact}
\end{table}

\subsubsection{Cross-Architecture Generalization}

\begin{table}[!ht]
\centering
\small

\begin{tabular}{lcccc}

\toprule
\textbf{Method}  & \textbf{\# Tokens} & \textbf{TextVQA} & \textbf{SQA} &  \textbf{MME} \\
\midrule
Vanilla &  1024 & 63.44 &  58.37 &  1702.84 \\
\hdashline
SVD-Prune (ours) & 900 & 62.84 &  57.37&  1716.38 \\
 &  784 & 61.80 &  58.94 &  1679.76 \\
 &  484 & 53.46 &  59.61 &  1650.04 \\
 &   256 & 37.88 &  57.14 &  1624.31 \\
\bottomrule
\end{tabular}
\caption{Performance of QwenVL with encoder-side SVD-Prune across different visual token budgets. }

\label{tab:qwenvl_svd}
\end{table}

Table~\ref{tab:qwenvl_svd} reports encoder-side pruning results on QwenVL. 
Despite this substantially higher token count, SVD-Prune remains effective under moderate compression. Retaining 784 tokens (23\% reduction) leads to only marginal degradation on TextVQA while slightly improving SQA over the vanilla model, suggesting that moderate pruning removes redundant visual information without harming reasoning performance. Under stronger compression, task sensitivity becomes more apparent: while SQA remains relatively stable and even improves at 484 retained tokens, TextVQA degrades significantly, indicating that text-centric tasks are more sensitive to aggressive token reduction. At 256 tokens (75\% reduction), this gap becomes more pronounced, with performance degradation driven primarily by the drop on TextVQA.


\subsection{Evaluation Across VLA}
\label{ssec:main_vla}

We next investigate whether the benefits of SVD-Prune extend beyond VLMs to embodied VLA settings. As reported in Table~\ref{tab:libero_svd_merged}, SVD-Prune maintains competitive action prediction performance while substantially reducing the visual token budget. On LIBERO-Spatial and LIBERO-Goal, success rates remain close to the unpruned baseline across the evaluated budgets, with some pruned configurations even yielding higher observed success rates. LIBERO-Object is similarly robust, with the success rate remaining within 1.4 points of the baseline even at 56 retained tokens. These results indicate that substantial visual token reduction can be applied to the three specialized LIBERO suites with limited degradation in task performance. LIBERO-10 is more sensitive to aggressive pruning, with success rate decreasing from 55.8\% at 256 tokens to 47.0\% at 56 tokens. This sensitivity may stem from its longer interaction horizons and multi-goal tasks, which require preserving relevant visual information across more complex action sequences.

\begin{figure*}[!ht]
    \centering
    \includegraphics[width=1.0\linewidth]{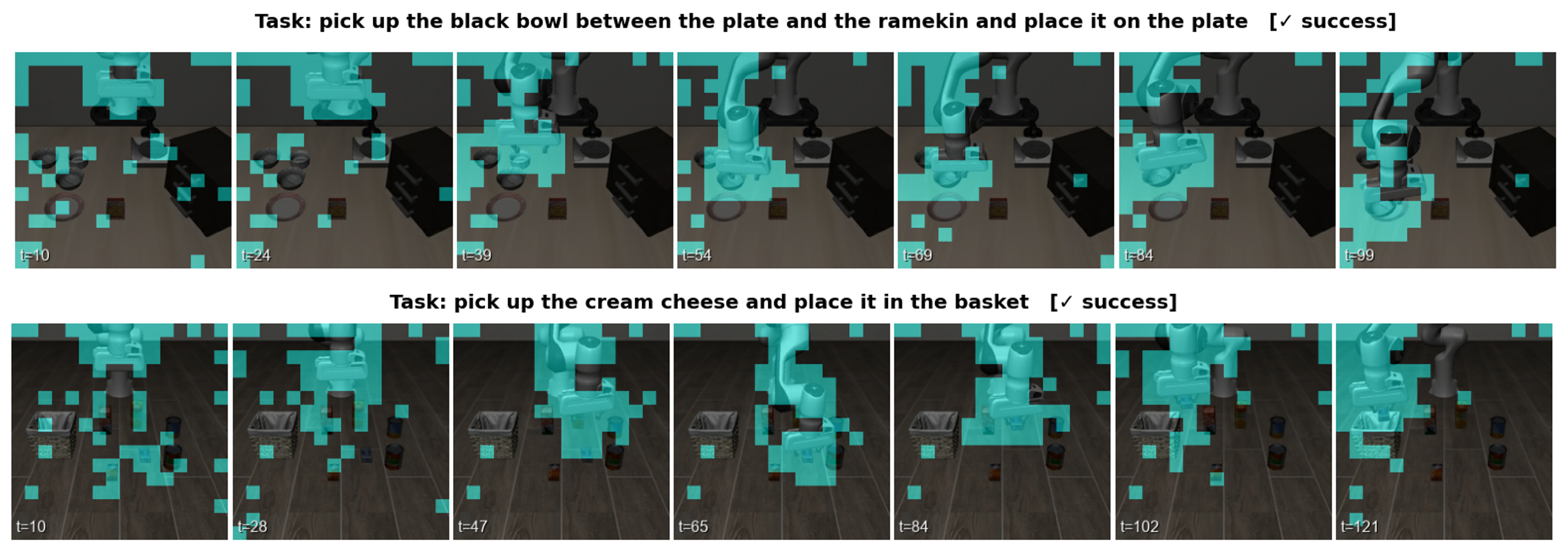}
    \caption{Qualitative visualization of SVD-Prune on LIBERO-Spatial (top) and LIBERO-Object (bottom), at selected frame indices. Input frames and corresponding decoder-side pruning results retaining 56 visual tokens are shown, with frame numbers indicated below each example.
    }
    \label{fig:vis_VLA}
\end{figure*}

Across all suites, the efficiency gains become increasingly pronounced as the token budget decreases. For example, on LIBERO-Spatial, reducing the budget from 256 to 56 tokens lowers computational cost from 4376.05 to 2349.89\,GFLOPs and inference latency from 1099.60 to 900.96\,ms. Figure~\ref{fig:vis_VLA} further illustrates how decoder-side SVD-Prune retains visual regions associated with manipulated objects across successive timesteps.

\begin{table*}[t]
\centering
\small
\setlength{\tabcolsep}{3.5pt} 
\begin{tabular}{@{}ccccccccccccc@{}}
\toprule
\multirow{2}{*}{\textbf{Tokens}} &
\multicolumn{3}{c}{\textbf{LIBERO-Spatial}} &
\multicolumn{3}{c}{\textbf{LIBERO-Goal}} &
\multicolumn{3}{c}{\textbf{LIBERO-Object}} &
\multicolumn{3}{c}{\textbf{LIBERO-10}} \\
\cmidrule(lr){2-4} \cmidrule(lr){5-7} \cmidrule(lr){8-10} \cmidrule(lr){11-13}
& \textbf{SR (\%)} & \textbf{GFLOPs} & \textbf{Lat. (ms)}
& \textbf{SR (\%)} & \textbf{GFLOPs} & \textbf{Lat. (ms)}
& \textbf{SR (\%)} & \textbf{GFLOPs} & \textbf{Lat. (ms)}
& \textbf{SR (\%)} & \textbf{GFLOPs} & \textbf{Lat. (ms)} \\
\midrule
256 
& 78.0 & 4376.05 & 1099.60 & 75.2 & 4213.82 & 1015.32 & 72.4 & 4254.36 & 1436.51 & 55.8 & 4294.92 & 3213.72 \\
\midrule
192            & 79.0 & 3935.92 & 1044.56 & 78.4 & 3774.09 & 1014.45 & 71.6 & 3814.53 & 1441.28 & 54.6 & 3854.89 & 2922.52 \\
128            & 80.8 & 3075.68 & 982.76  & 76.8 & 3336.51 & 983.44  & 73.0 & 3376.85 & 1399.32 & 51.8 & 3417.31 & 2631.32 \\
56             & 79.6 & 2349.89 & 900.96  & 79.6 & 2846.80 & 988.74  & 71.0 & 2887.03 & 988.13  & 47.0 & 2927.27 & 2304.10 \\
\bottomrule
\end{tabular}
\caption{Task success rate and inference efficiency of OpenVLA with SVD-Prune on LIBERO-Spatial, LIBERO-Goal, LIBERO-Object, and LIBERO-10. Each suite comprises 10 tasks with 50 trials per task. Decoder-side pruning is applied after the 16th decoder layer.}
\label{tab:libero_svd_merged}
\end{table*}

\section{Ablation Studies}

\subsection{Energy threshold}
\label{sec:eps_ablation}

To determine a suitable spectral energy threshold $\varepsilon$, we conduct a sensitivity analysis on three benchmarks across multiple token budgets (16 and 192 tokens). As shown in Table~\ref{tab:epsilon_ablation}, performance remains relatively low and stable for $\varepsilon \in [0.1, 0.7]$, indicating that retaining only a small fraction of the spectral energy is insufficient to preserve task-relevant visual structure. Performance improves markedly once $\varepsilon \geq 0.75$, with accuracy increasing consistently across token budgets and reaching 59.9\% and 52.8\% on GQA for the 192- and 16-token settings, respectively, at $\varepsilon = 0.8$. This transition suggests the presence of a critical spectral-energy threshold beyond which the retained subspace becomes sufficiently expressive to preserve the dominant representation structure required for downstream reasoning. Further increasing $\varepsilon$ beyond 0.8 yields only marginal gains, with near-saturated performance for $\varepsilon \in [0.8, 0.9]$ and slight fluctuations at $\varepsilon = 0.95$, particularly in low-token regimes. Based on this trade-off between performance and computational overhead, we use $\varepsilon = 0.8$ as the default setting in all experiments.

\begin{table}
\centering
\small
\begin{tabular}{@{}c cc cc cc@{}}
\toprule
& \multicolumn{2}{c}{GQA}~\cite{hudson2019gqa} & \multicolumn{2}{c}{POPE}~\cite{li2023pope} & \multicolumn{2}{c}{MME}~\cite{fu2023mme} \\
\cmidrule(lr){2-3} \cmidrule(lr){4-5} \cmidrule(lr){6-7}
$\varepsilon$ & 192 &  16 & 192 & 16 & 192 & 16 \\
\midrule
0.10 & 38.30 & 38.30 & 50.53 & 48.93 & 1123.71 & 1151.58 \\
0.70 & 40.61 & 38.67 & 52.22 & 52.80 & 1150.03 & 1167.41 \\
0.75 & 45.95 & 40.87 & 86.30 & 71.91 & 1579.33 & 1283.55 \\
\hdashline
0.80 & 59.88 & 52.80  & 87.75 & 72.68 & 1788.00 & 1364.13 \\
\hdashline
0.85 & 59.90 & 52.86 & 87.76 & 72.77 & 1789.01 & 1364.24 \\
0.90 & 59.90 & 53.00 & 88.22 & 72.60 &  1788.97 & 1364.24 \\
0.95 & 60.05 & 52.60 & 88.00 & 72.60 &1789.00 & 1364.31 \\
\bottomrule
\end{tabular}
\caption{Effect of the spectral energy retention threshold $\varepsilon$ on the performance of LLaVA-1.5-7B across multiple benchmarks and token budgets.}
\label{tab:epsilon_ablation}

\end{table}

\subsection{Scoring Function}
\label{ssec:Scoring}

We compare the standard leverage-score criterion used in SVD-Prune against energy-weighted~\cite{sun2024energy}, whitening-based~\cite{kessy2018optimal}, and Augmented Visual Importance (AVI) scoring. Energy weighting emphasizes dominant spectral directions, whitening favors lower-energy directions, while AVI combines contributions from the left and right singular subspaces. Table~\ref{tab:scoring} shows that standard leverage scoring is the most robust criterion across the evaluated benchmarks. Reweighting the singular spectrum through energy weighting or whitening substantially degrades TextVQA performance, while AVI partially recovers accuracy but remains below leverage scoring. This suggests that the original leverage formulation already provides an effective token-importance ranking, motivating its use as the default criterion in SVD-Prune.

\begin{table}[t]
\centering
\small
\setlength{\tabcolsep}{12pt}
\begin{tabular}{lccc}
\toprule
\textbf{Scoring function} & \textbf{TextVQA}
& \textbf{POPE}
& \textbf{MME}
\\
\midrule
Energy-weighted     & 42.07  & 76.16 & 1204.48\\
Whitening-based     & 42.00  & 83.16 & 1536.33 \\
AVI                 & 53.23  & 83.68 & 1539.07 \\
Leverage     & \textbf{55.14}  & \textbf{83.87} & \textbf{1575.00}\\
\bottomrule
\end{tabular}
\caption{Ablation of token scoring functions on LLaVA-1.5-7B with encoder-side pruning and a fixed budget of 64 visual tokens.}
\label{tab:scoring}
\end{table}

\section{Conclusion}
\label{sec:conclusion}

We introduced SVD-Prune, a training-free and architecture-agnostic vision token pruning framework that formulates token reduction as a low-rank subspace preservation problem. By leveraging spectral decomposition and leverage-score-based token selection, SVD-Prune provides a globally informed alternative to conventional local token scoring strategies. Extensive experiments across VLMs and VLA models demonstrate that preserving the dominant spectral structure enables robust token pruning even under extreme compression. In particular, SVD-Prune remains highly effective in ultra-low-token regimes, retaining competitive multimodal reasoning and action prediction performance while reducing the visual token budget by up to 99.3\%, inference FLOPs by 84.8\%, and end-to-end latency by 4.18$\times$. At a matched 16-token budget, SVD-Prune also provides lower latency than FastV and SparseVLM, while its SVD overhead remains small relative to the overall model cost. On OpenVLA, retaining only 56 of the original 256 visual tokens (78\% reduction) maintains near-baseline success rates on three of the four LIBERO suites.

Our analysis further reveals that pruning effectiveness depends on feature contextualization, pruning location, and task characteristics. While reasoning-oriented tasks and binary yes/no queries remain robust even in ultra-low-token regimes, OCR-intensive tasks are more sensitive to pruning location and depth. Finally, the lack of explicit diversity constraints in leverage-based selection motivates future investigation of diversity-aware strategies.\\
\\
\textbf{Acknowledgments}\\
This work was supported by the European Union's Horizon Europe research and innovation programme under the Marie Skłodowska-Curie COFUND grant agreement No 101127936 (DeMythif.AI), and France 2030 funding managed by the National Research Agency (ANR), as part of IA CLUSTER program, reference ANR-23-IACL-0003-DATAIA CLUSTER. \\

\textbf{Use of AI Tools.} The authors used AI tools solely for language editing. All scientific ideas, content, and results were developed and verified by the authors.



\clearpage
\onecolumn

\newcount\supppage
\newcount\supppages

\pdfximage{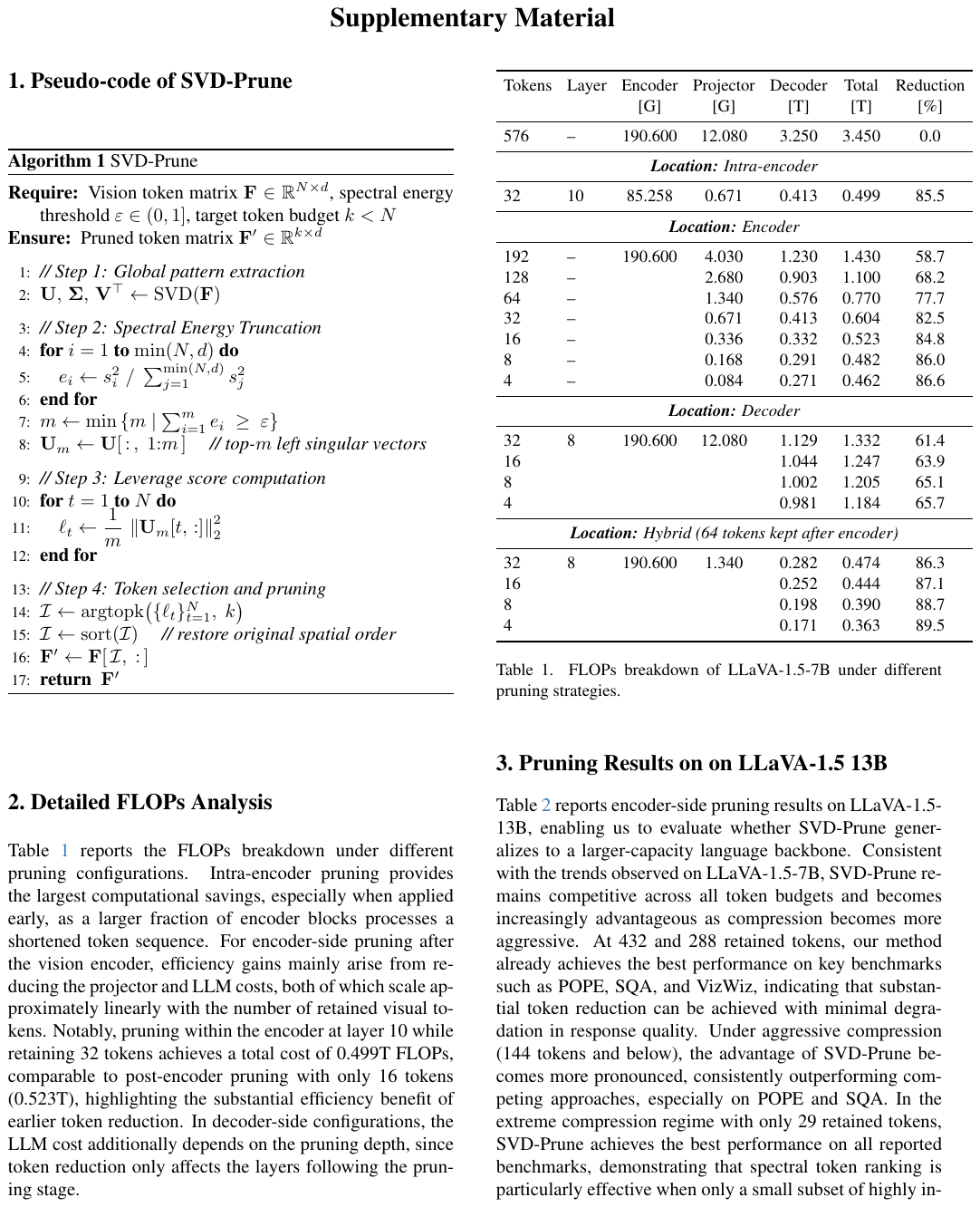}
\supppages=\pdflastximagepages
\supppage=0

\loop
    \advance\supppage by 1
    \thispagestyle{empty}
    \edef\suppCurrentPage{\the\supppage}

    \AddToShipoutPicture*{%
        \AtPageLowerLeft{%
            \includegraphics[
                page=\suppCurrentPage,
                width=\paperwidth,
                height=\paperheight,
                keepaspectratio
            ]{SVD_Prune_WACV_Supplementary_materials.pdf}%
        }%
    }%

    \null\clearpage
\ifnum\supppage<\supppages
\repeat


\end{document}